# Dynamic models for Planar Peristaltic Locomotion of a Metameric Earthworm-like Robot

**Qinyan Zhou[1], Hongbin Fang[1, §], Zhihai Bi[1], Jian Xu[1]**

[1)] Institute of AI and Robotics, MOE Engineering Research Center of AI & Robotics, State Key Laboratory of Medical Neurobiology, Fudan University, Shanghai 20043, China

[§] **Corresponding author:** fanghongbin@fudan.edu.cn **(H. Fang)**

**Abstract:**

The development of versatile robots capable of traversing challenging and irregular environments is of increasing interest in the field of robotics, and metameric robots have been identified as a promising solution due to their slender, deformable bodies. Inspired by the effective locomotion of earthworms, earthworm-like robots capable of both rectilinear and planar locomotion have been designed and prototyped. While much research has focused on developing kinematic models to describe the planar locomotion of earthworm-like robots, the authors argue that the development of dynamic models is critical to improving the accuracy and efficiency of these robots. A comprehensive analysis of the dynamics of a metemeric earthworm-like robot capable of planar motion is presented in this work. The model takes into account the complex interactions between the robot's deformable body and the forces acting on it, and draws on the methods previously used to develop mathematical models of snake-like robots. The proposed model represents a significant advancement in the field of metameric robotics and has the potential to enhance the performance of earthworm-like robots in a variety of challenging environments, such as underground pipes and tunnels, and serves as a foundation for future research into the dynamics of soft-bodied robots.

**Keywords:** Metameric robots, Earthworm-like robots, Dynamic modeling, Planar locomotion, Mathematical models

# 1 Introduction

Recent years have seen an increasing need and interest in developing locomotion robots due to the urgent requirement to liberate workers from tasks in rugged and hazardous environments. For confined work spaces and irregular work environments, traditional wheeled and legged locomotion robots face major limitations such as inefficient movement and low adaptability to continuously changing environments [1,2]. Instead, metameric robots enable readily traveling within unstructured environments and versatility in diverse working conditions due to their slender bodies and deformable components [3–6].

In particular, many metameric robots are inspired by the earthworms [7–11] for their effective locomotion in a variety of environments, especially in confined spaces such as underground caves. The slender body of earthworms consists of many deformable segments that can be stretched, contracted, and curved by control of embedded longitudinal and circumferential muscles [12,13]. The robotics field has greatly benefited from mimicking the locomotion of the earthworm, with a wealth of new robotic designs and prototypes being proposed for scenarios such as earthquake rescue, industrial pipeline inspection, and terrain exploration [14–16].

For common earthworm-like robots that perform 1-D rectilinear motions in pipes, many models have been developed by mimicking the segmentation of the earthworm and reproducing the retrograde peristalsis waves[8,9,17,18]. In this regard, some kinematic and dynamic models describing the rectilinear locomotion of the earthworm-like robot have been proposed to evaluate the locomotion performance of the robot and to further optimize the gaits [19–23]. Other than rectilinear locomotion, earthworms in nature are capable of effective planar locomotion. By differentially contracting or stretching the longitudinal muscles on the left and right sides, the earthworm can bend its body and turn [13]. Motivated by this observation, several earthworm-like robots capable of planar locomotion have been designed and prototyped, the main idea being to make the left and right sides of the segment unequally deformed[10,16,24]. Many body segments and various segmental states permit earthworm-like robots with good mobility and abundant locomotion gaits. Despite the progresses in achieving effective planar locomotion of earthworm-like robots, the theoretical descriptions of planar locomotion are currently largely resting on kinematic models, and the resulting discrepancies between theory and experiment necessitate further efforts in the dynamic modeling of the planar locomotion of the earthworm-like robots [10,25].

In addition, due to the similarity between earthworm-like robots and snake-like robots in terms of planar locomotion, one line of thought is to learn the dynamics model of the earthworm-like robot from the snake-like robot. The mathematical models of planar snake robot dynamics have previously been developed [26,27], and the initial form of the model presented in this work is developed using the same approach as in these works.

# 2 Robot Modeling

## 2.1 Kinematic modeling

**Table 1** Parameters that characterize the earthworm-like robot

| Symbol | Description | Vector |
|---|---|---|
| $n$ | The number of segments | |
| $r$ | Radius of the rigid body | |
| $m$ | Mass of each rigid body | |
| $J$ | Moment of inertia of each rigid body | |
| $(x_i, y_i)$ | Global coordinates of the CM (center of mass) of the rigid body $i$ | $\mathbf{X}, \mathbf{Y} \in \mathbb{R}^{n+1}$ |
| $(l_{i,x}, l_{i,y})$ | The distance between the CM of the rigid body $i$ and the rigid body $i+1$ in the global coordinate system | $\mathbf{L_x}, \mathbf{L_y} \in \mathbb{R}^{n}$ |
| $(l_{i,l}, l_{i,r})$ | The left (right) length between the rigid body $i$ and the rigid body $i+1$ | |
| $\theta_i$ | The angle between the normal direction of the rigid body $i$ and the positive $x_i$ direction | $\boldsymbol{\theta} \in \mathbb{R}^{n+1}$ |
| $\varphi_i$ | The angle between the rigid body $i$ and the rigid body $i+1$ | |
| $(f_{i,local,x}, f_{i,local,y})$ | Coulomb friction force on the rigid body $i$ in the local frame | |
| $(f_{i,global,x}, f_{i,global,y})$ | Coulomb friction force on the rigid body $i$ | $\mathbf{f_x}, \mathbf{f_y} \in \mathbb{R}^{n+1}$ |
| $(h_{i,l}, h_{i,r})$ | The inner forces generated by the left (right) actuator of segment $i$ | |
| $(h_{i,l,x}, h_{i,r,x}), (h_{i,l,y}, h_{i,r,y})$ | The inner forces generated by the left (right) actuator of segment $i$ in the global coordinate system | $\mathbf{H_x}, \mathbf{H_y} \in \mathbb{R}^{n}$ |
| $\xi_{tp}$, $\xi_{tn}$ and $\xi_n$ | The coefficients describing the Coulomb friction force in the tangential forward (along the positive direction of the body $x$ axis), tangential backward (along the negative direction of the body $x$ axis) , and normal (along the body $y$ axis) directions of a rigid body | |

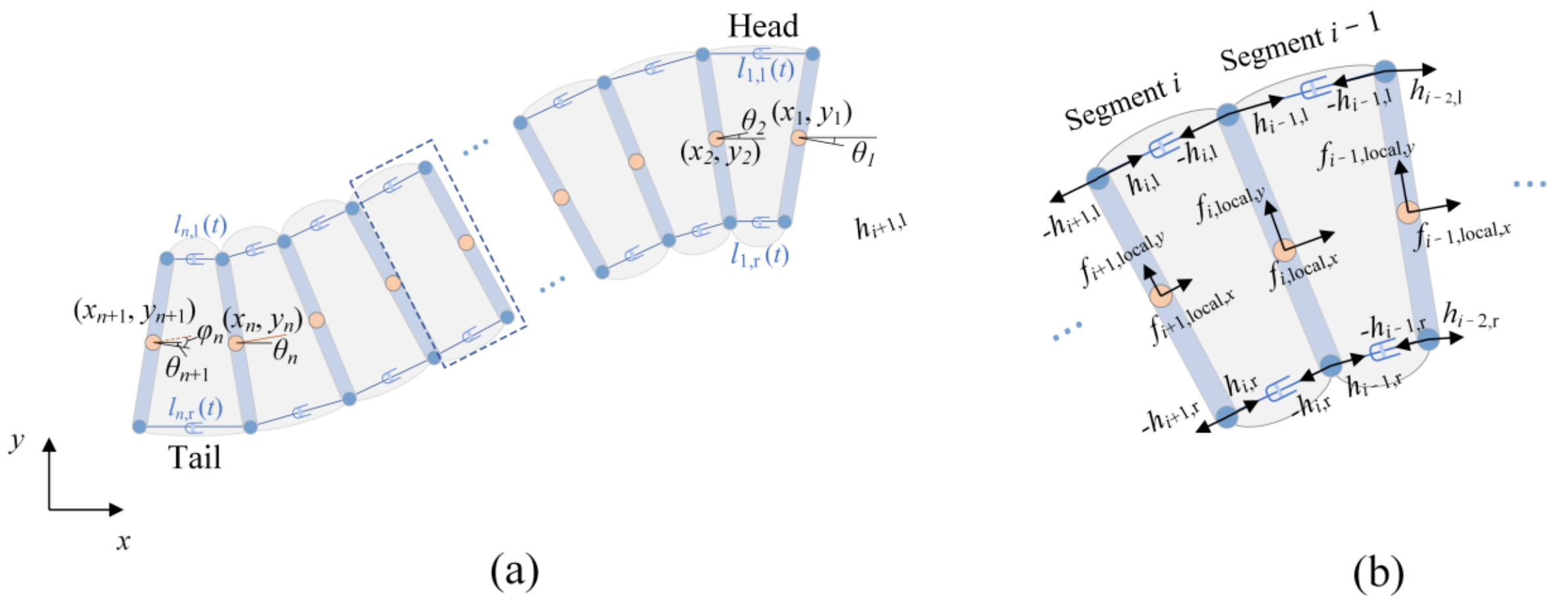

**FIG. 1.** Mechanical model of the metameric robot. (a) An $n$-segment model of the robot consists of $n$ rigid bodies interconnected by $n$+1 pairs of displacement actuators. (b) Free-body diagrams of segment $i$ and segment $i$-1, with orange circles representing the center of mass and arrows indicating forces.

By establishing a bionic correspondence between the earthworm's morphological characteristics and the robot's mechanical parts, a earthworm-like module is designed (Fig.1a). Specifically, the worm-like module is made up of two acrylic plates (referred to as the front and the rear plate, respectively), two servomotors, eight spring-steel belts, servomotor-driven cords, and bristle structures. In the following subsections, the kinematics and dynamics of the snake robot will be modelled in terms of the mathematical symbols described in Table 1.

We will make use of the following vectors and matrices:

$$\mathbf{D}_1 = \begin{bmatrix} -1 & 1 & & & \\ & -1 & 1 & & \\ & & \cdots & & \\ & & & -1 & 1 \end{bmatrix} \in \mathbb{R}^{N\times(N+1)}, \qquad \mathbf{e} = [1,\ldots,1]^T \in \mathbb{R}^{N+1},$$

$$\mathbf{D}_2 = \begin{bmatrix} -1 & -1 & & & & & & & & \\ 1 & 1 & -1 & -1 & & & & & & \\ & & 1 & 1 & -1 & -1 & & & & \\ & & & & & \cdots & & & & \\ & & & & & & -1 & -1 & & \\ & & & & & & 1 & 1 & -1 & -1 \\ & & & & & & & & 1 & 1 \end{bmatrix} \in \mathbb{R}^{(N+1)\times 2N},$$

$$\mathbf{D}_3 = \begin{bmatrix} 1 & -1 & & & & & & & & \\ -1 & 1 & 1 & -1 & & & & & & \\ & & -1 & 1 & 1 & -1 & & & & \\ & & & & & \cdots & & & & \\ & & & & & & 1 & -1 & & \\ & & & & & & -1 & 1 & 1 & -1 \\ & & & & & & & & -1 & 1 \end{bmatrix} \in \mathbb{R}^{(N+1)\times 2N},$$

$\mathbf{X} = \left[x_1, \dots, x_{N+1}\right]^T \in \mathbb{R}^{N+1}$, $\mathbf{Y} = \left[y_1, \dots, y_{N+1}\right]^T \in \mathbb{R}^{N+1}$, $\boldsymbol{\theta} = \left[\theta_1, \dots, \theta_{N+1}\right]^T \in \mathbb{R}^{N+1}$,

$\sin\boldsymbol{\theta} = \left[\sin\theta_1, \dots, \sin\theta_{N+1}\right]^T \in \mathbb{R}^{N+1}$, $\cos\boldsymbol{\theta} = \left[\cos\theta_1, \dots, \cos\theta_{N+1}\right]^T \in \mathbb{R}^{N+1}$,

$\mathbf{S}_\theta = \mathrm{diag}\left(\sin\boldsymbol{\theta}\right) \in \mathbb{R}^{(N+1)\times(N+1)}$, $\mathbf{C}_\theta = \mathrm{diag}\left(\cos\boldsymbol{\theta}\right) \in \mathbb{R}^{(N+1)\times(N+1)}$, $\mathbf{f}_x = \left[f_{1,x}, \dots, f_{N+1,x}\right]^T \in \mathbb{R}^{N+1}$,

$\mathbf{f}_y = \left[f_{1,y}, \dots, f_{N+1,y}\right]^T \in \mathbb{R}^{N+1}$, $\mathbf{R}_{local,i}^{\mathrm{global}} = \begin{bmatrix} \cos\theta_i & -\sin\theta_i \\ \sin\theta_i & \cos\theta_i \end{bmatrix}$,

$\mathbf{H}_x = \left[h_{1,l,x}, h_{1,r,x}, \dots, h_{i-1,l,x}, h_{i-1,r,x}, h_{i,l,x}, h_{i,r,x}, \dots, h_{N,l,x}, h_{N,r,x}\right]^T \in \mathbb{R}^{2N}$,

$\mathbf{H}_y = \left[h_{1,l,y}, h_{1,r,y}, \dots, h_{i-1,l,y}, h_{i-1,r,y}, h_{i,l,y}, h_{i,r,y}, \dots, h_{N,l,y}, h_{N,r,y}\right]^T \in \mathbb{R}^{2N}$.

In the proposed model of the earthworm-like robot (Figure 1), each segment (denoted by a dashed box) consists of a pair of idealized massless displacement actuator and a rigid body of mass $m$ and moment of inertia $J = ml^2/3$ located on the right. Figure 1a displays an $n$ -segment model of the metameric robot. To satisfy kinematics, an additional rigid body is added at the back end of the model. Hence, the $n$-segment model consists of $n+1$ rigid bodies interconnected by $n$ pairs of displacement actuators.

For each two adjacent rigid bodies, the distance of the left and right ends are accurately and independently controlled by an ideal displacement actuator, respectively. Assumed that the contraction directions on both sides are parallel to each other and form an isosceles trapezoid with the two rigid bodies as illustrated in Figure 1, the translation and rotation between adjacent rigid bodies following periodic law with respect to time:

$$x_i - x_{i+1} = l_{i,x} = \left(l_{i,l} + l_{i,r}\right)\cos\left(\theta_{i+1} + \varphi_i/2\right)/2, \tag{1.a}$$

$$y_i - y_{i+1} = l_{i,y} = \left(l_{i,l} + l_{i,r}\right)\sin\left(\theta_{i+1} + \varphi_i/2\right)/2, \tag{1.b}$$

$$\theta_i - \theta_{i+1} = \varphi_i = \arccos\left[\frac{8r^2 - \left(l_{i,l} - l_{i,r}\right)^2}{8r^2}\right] \cdot \mathrm{sgn}\left(l_{i,l} - l_{i,r}\right). \tag{1.c}$$

We can write the translation constraints (1.a) and (1.b) for all the segments of the robot in matrix form as

$$\mathbf{D}_1\mathbf{X} = \mathbf{L}_x, \tag{2.a}$$

$$\mathbf{D}_1\mathbf{Y} = \mathbf{L}_y. \tag{2.b}$$

The rigid body accelerations may also be expressed by differentiating (2.a) and (2.b) twice with respect to time, which gives

$$\mathbf{D}_1\ddot{\mathbf{X}} = \ddot{\mathbf{L}}_x, \tag{3.a}$$

$$\mathbf{D}_1\ddot{\mathbf{Y}} = \ddot{\mathbf{L}}_y. \tag{3.b}$$

**2.2 Dynamics modeling**

The forces determining the translation and rotation of the robot in the *xoy* plane are shown in Figure **1b**. By employing Newton's second law and the theorem of angular momentum, we have

$$m\ddot{x}_i = f_{i,x} + h_{i-1,l,x} + h_{i-1,r,x} - h_{i,l,x} - h_{i,r,x}, \tag{4.a}$$

$$m\ddot{y}_i = f_{i,y} + h_{i-1,l,y} + h_{i-1,r,y} - h_{i,l,y} - h_{i,r,y}. \tag{4.b}$$

The force balance equations for all rigid bodies may be expressed in matrix form as

$$m\ddot{\mathbf{X}} = \mathbf{f}_x + \mathbf{D}_2\mathbf{H}_x, \tag{5.a}$$

$$m\ddot{\mathbf{Y}} = \mathbf{f}_y + \mathbf{D}_2\mathbf{H}_y. \tag{5.b}$$

The torque balance for rigid body $i$ is given by

$$\begin{aligned} J\ddot{\theta}_i = &\frac{r}{2}\cos\theta_i\left(-h_{i-1,l,x} + h_{i-1,r,x} + h_{i,l,x} - h_{i,r,x}\right) \\ &+\frac{r}{2}\sin\theta_i\left(-h_{i-1,l,y} + h_{i-1,r,y} + h_{i,l,y} - h_{i,r,y}\right), \end{aligned} \tag{6}$$

and the torque balance equations for all links may be expressed in matrix form as

$$J\ddot{\boldsymbol{\theta}} = \frac{r}{2}\mathbf{C}_\theta\mathbf{D}_3\mathbf{H}_x + \frac{r}{2}\mathbf{S}_\theta\mathbf{D}_3\mathbf{H}_y. \tag{7}$$

It now remains to remove the joint constraint forces from(7). Premultiplying (5.a) and (5.b) by $\mathbf{D}_1$, solving for $\mathbf{H}_x$ and $\mathbf{H}_y$, and also inserting (3.a) and (3.b) give

$$\mathbf{H}_x = \left(\mathbf{D}_1\mathbf{D}_2\right)^{-1}\left(m\ddot{\mathbf{L}}_x - \mathbf{D}_1\mathbf{f}_x\right), \tag{8.a}$$

$$\mathbf{H}_y = \left(\mathbf{D}_1\mathbf{D}_2\right)^{-1}\left(m\ddot{\mathbf{L}}_y - \mathbf{D}_1\mathbf{f}_y\right). \tag{8.b}$$

By inserting (8.a) and (8.b) into (7) and solving for $\ddot{\boldsymbol{\theta}}$, we can finally rewrite the model of the snake robot as

$$\begin{aligned} J\ddot{\boldsymbol{\theta}} = &\frac{r}{2}\mathbf{C}_\theta\mathbf{D}_3\left(\mathbf{D}_1\mathbf{D}_2\right)^{-1}\left(m\ddot{\mathbf{L}}_x - \mathbf{D}_1\mathbf{f}_x\right) \\ &+\frac{r}{2}\mathbf{S}_\theta\mathbf{D}_3\left(\mathbf{D}_1\mathbf{D}_2\right)^{-1}\left(m\ddot{\mathbf{L}}_y - \mathbf{D}_1\mathbf{f}_y\right). \end{aligned} \tag{9}$$

In this research, the head displacement $x_1$ and the head velocity $\dot{x}_1$ are respectively employed to characterize the displacement and velocity of the robot as a whole. The linear displacement and the angular displacement of each rigid body can be expressed in terms of the displacement of the head body according to (1.a), (1.b) and (1.c), which gives

$$x_i = x_1 - \sum_{j=1}^{i-1} l_{i,x}, \tag{10.a}$$

$$y_i = y_1 - \sum_{j=1}^{i-1} l_{i,y}, \tag{10.b}$$

$$\theta_i = \theta_1 - \sum_{j=1}^{i-1} \varphi_i. \tag{10.c}$$

Hence, differentiating (10.a), (10.b) and (10.c) twice with respect to time and summing up the acceleration of all rigid bodies gives

$$\sum_{i=1}^{n+1} \ddot{x}_i = (n+1)\ddot{x}_1 - \sum_{i=1}^{n+1}\sum_{j=1}^{i-1} \ddot{l}_{j,x}, \tag{11.a}$$

$$\sum_{i=1}^{n+1} \ddot{y}_i = (n+1)\ddot{y}_1 - \sum_{i=1}^{n+1}\sum_{j=1}^{i-1} \ddot{l}_{j,y}, \tag{11.b}$$

$$\sum_{i=1}^{n+1} \ddot{\theta}_i = (n+1)\ddot{\theta}_1 - \sum_{i=1}^{n+1}\sum_{j=1}^{i-1} \ddot{\varphi}_j. \tag{11.c}$$

Hence, the linear acceleration of the robot gives by inserting (11.a) and (11.b) into (5.a) and (5.b), and noting that the inner forces, $\mathbf{H}_x$ and $\mathbf{H}_y$, are eliminated when the link accelerations are summed (i.e. $\mathbf{e}^T\mathbf{D}_2 = \mathbf{0}$). This gives

$$\begin{aligned} & m\left[(n+1)\ddot{x}_1 - \sum_{i=1}^{n+1}\sum_{j=1}^{i-1} \ddot{l}_{j,x}\right] = \sum_{i=1}^{n+1} f_{i,x}, \\ & \ddot{x}_1 = \frac{1}{n+1}\left(\frac{1}{m}\sum_{i=1}^{n+1} f_{i,x} + \sum_{i=1}^{n+1}\sum_{j=1}^{i-1} \ddot{l}_{j,x}\right), \end{aligned} \tag{12.a}$$

$$\begin{aligned} & m\left[(n+1)\ddot{y}_1 - \sum_{i=1}^{n+1}\sum_{j=1}^{i-1} \ddot{l}_{j,y}\right] = \sum_{i=1}^{n+1} f_{i,y}, \\ & \ddot{y}_1 = \frac{1}{n+1}\left(\frac{1}{m}\sum_{i=1}^{n+1} f_{i,y} + \sum_{i=1}^{n+1}\sum_{j=1}^{i-1} \ddot{l}_{j,y}\right). \end{aligned} \tag{12.b}$$

Similarly, the angular acceleration of the robot can be obtained by inserting (11.c) into (9),

$$\ddot{\theta}_1 = \frac{1}{n+1}\left\{\frac{1}{J}\begin{bmatrix} \frac{r}{2}\mathbf{e}^T\mathbf{C}_\theta\mathbf{D}_3\left(\mathbf{D}_1\mathbf{D}_2\right)^{-1}\left(m\ddot{\mathbf{L}}_x - \mathbf{D}_1\mathbf{f}_x\right) \\ +\frac{r}{2}\mathbf{e}^T\mathbf{S}_\theta\mathbf{D}_3\left(\mathbf{D}_1\mathbf{D}_2\right)^{-1}\left(m\ddot{\mathbf{L}}_y - \mathbf{D}_1\mathbf{f}_y\right) \end{bmatrix} + \sum_{i=1}^{n+1}\sum_{j=1}^{i-1}\ddot{\varphi}_j\right\}. \tag{13}$$

The local coordinate system of each link is fixed in the center of mass of the link with $x$ (tangential) and $y$ (normal) axes oriented such that they are aligned with the global $x$ and $y$ axis, respectively, when $\theta_i = 0$. The rotation matrix from the global frame to the frame of rigid body $i$ is given by

$$\mathbf{R}_{local,i}^{\text{global}} = \begin{bmatrix} \cos\theta_i & -\sin\theta_i \\ \sin\theta_i & \cos\theta_i \end{bmatrix}. \tag{14}$$

To provide anisotropic friction, the brush structure is attached to the bottom of the segment. The brushes are tilted back $45^{\circ}$ with very sharp side edges so that the robotic segment is subjected to little friction when pulled/pushed forward, but can grip rough surfaces when moving backward or laterally. The coefficients describing the Coulomb friction force in the tangential forward, tangential backward and normal directions of a rigid body, respectively, are denoted by $\xi_{tp}$, $\xi_{tn}$ and $\xi_n$, respectively. We define the Coulomb friction force on body $i$ in the local link frame,

$$\begin{bmatrix} f_{i,local,x} \\ f_{i,local,y} \end{bmatrix} = -mg \begin{bmatrix} \xi_t & 0 \\ 0 & \xi_n \end{bmatrix} \text{sgn}\left( \left(\mathbf{R}_{local,i}^{\text{global}}\right)^T \begin{bmatrix} \dot{x}_i \\ \dot{y}_i \end{bmatrix} \right), \tag{15.a}$$

$$\xi_t = \begin{cases} \xi_{t,+}, \text{if } \dot{x}_i \cos\theta_i + \dot{y}_i \sin\theta_i > 0 \\ \xi_{t,-}, \text{if } \dot{x}_i \cos\theta_i + \dot{y}_i \sin\theta_i < 0 \end{cases}. \tag{15.b}$$

Where $\left(\mathbf{R}_{local,i}^{\text{global}}\right)^T \begin{bmatrix} \dot{x}_i & \dot{y}_i \end{bmatrix}^T$ is the body velocity expressed in the local frame of rigid body $i$, and $g$ is the gravitational acceleration constant. Using the rotation matrix, we can express the global frame Coulomb friction force on rigid body $i$ as

$$\begin{bmatrix} f_{i,global,x} \\ f_{i,global,y} \end{bmatrix} = -mg\mathbf{R}_{local,i}^{\text{global}} \begin{bmatrix} \xi_t & 0 \\ 0 & \xi_n \end{bmatrix} \text{sgn}\left( \left(\mathbf{R}_{local,i}^{\text{global}}\right)^T \begin{bmatrix} \dot{x}_i \\ \dot{y}_i \end{bmatrix} \right). \tag{16}$$

By assembling the forces on all rigid bodies in matrix form, we can rewrite the global frame Coulomb friction forces on the links as

$$\begin{bmatrix} \mathbf{f}_x \\ \mathbf{f}_y \end{bmatrix} = -mg \begin{bmatrix} \xi_t \mathbf{C}_\theta & -\xi_n \mathbf{S}_\theta \\ \xi_t \mathbf{S}_\theta & \xi_n \mathbf{C}_\theta \end{bmatrix} \text{sgn}\left( \begin{bmatrix} \mathbf{C}_\theta & \mathbf{S}_\theta \\ -\mathbf{S}_\theta & \mathbf{C}_\theta \end{bmatrix} \begin{bmatrix} \dot{\mathbf{X}} \\ \dot{\mathbf{Y}} \end{bmatrix} \right). \tag{17}$$

## 3 Experimental verifications

In this section, we use the six-segment earthworm-like robot to verify the predicted locomotion modes and the corresponding kinematics characteristics. Here, we select some typical gaits which correspond to rectilinear, sidewinding, and circular locomotion, respectively to verify the locomotion characteristics. Figures 2, 3, and 4 show the actuation signals and experimental data of the robot for the three modes, and for reference, the theoretical trajectories are given in the insets.

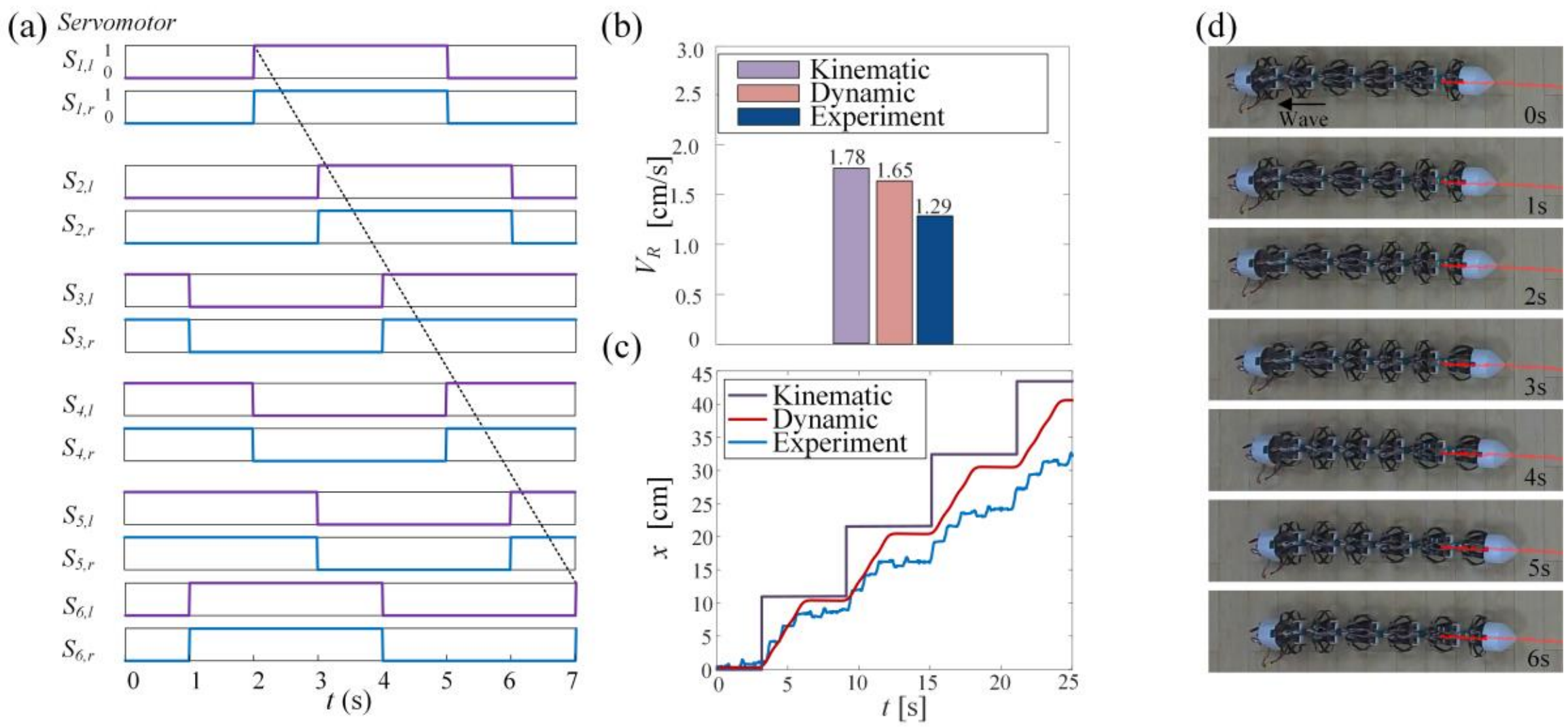

**FIG. 2.** Actuation signals and experiment of the earthworm-like rectilinear locomotion. (a) Actuation signals of the 6 robot segments in a locomotion cycle. (b) Comparison of the kinematic, dynamic, and experimental average velocities of the rectilinear locomotion. (c) Displacement-time histories of the robot head under the rectilinear gait in 25 seconds. (d) Snapshots of the robot during the rectilinear locomotion.

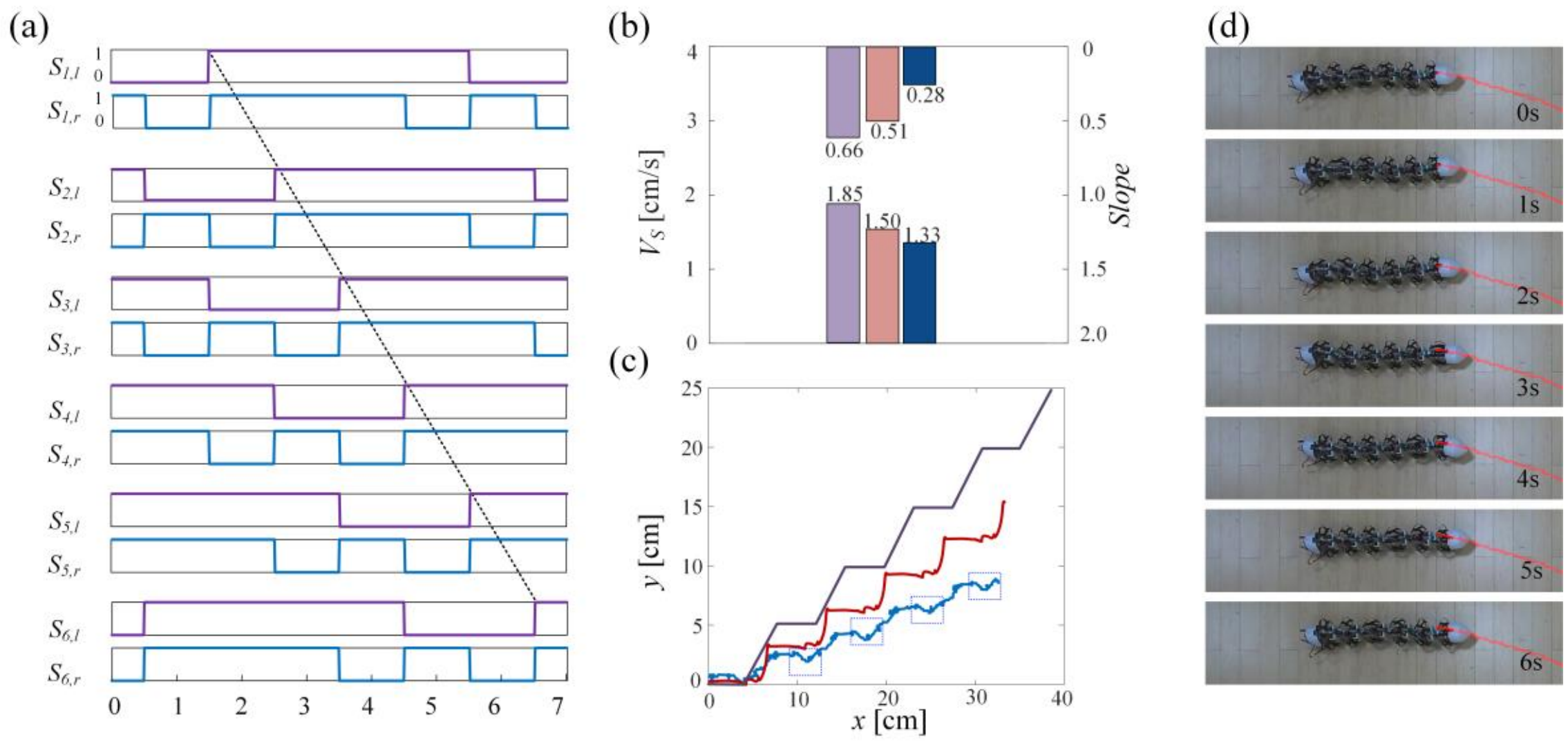

**FIG. 3.** Actuation signals and experiment of the earthworm-like sidewinding locomotion. (a) Actuation signals of the 6 robot segments in a locomotion cycle. (b) Comparison of the kinematic, dynamic, and experimental average velocities and slope of the sidewinding locomotion. (c) Displacement-time histories of the robot head under the sidewinding gait. (d) Snapshots of the robot during the sidewinding locomotion.

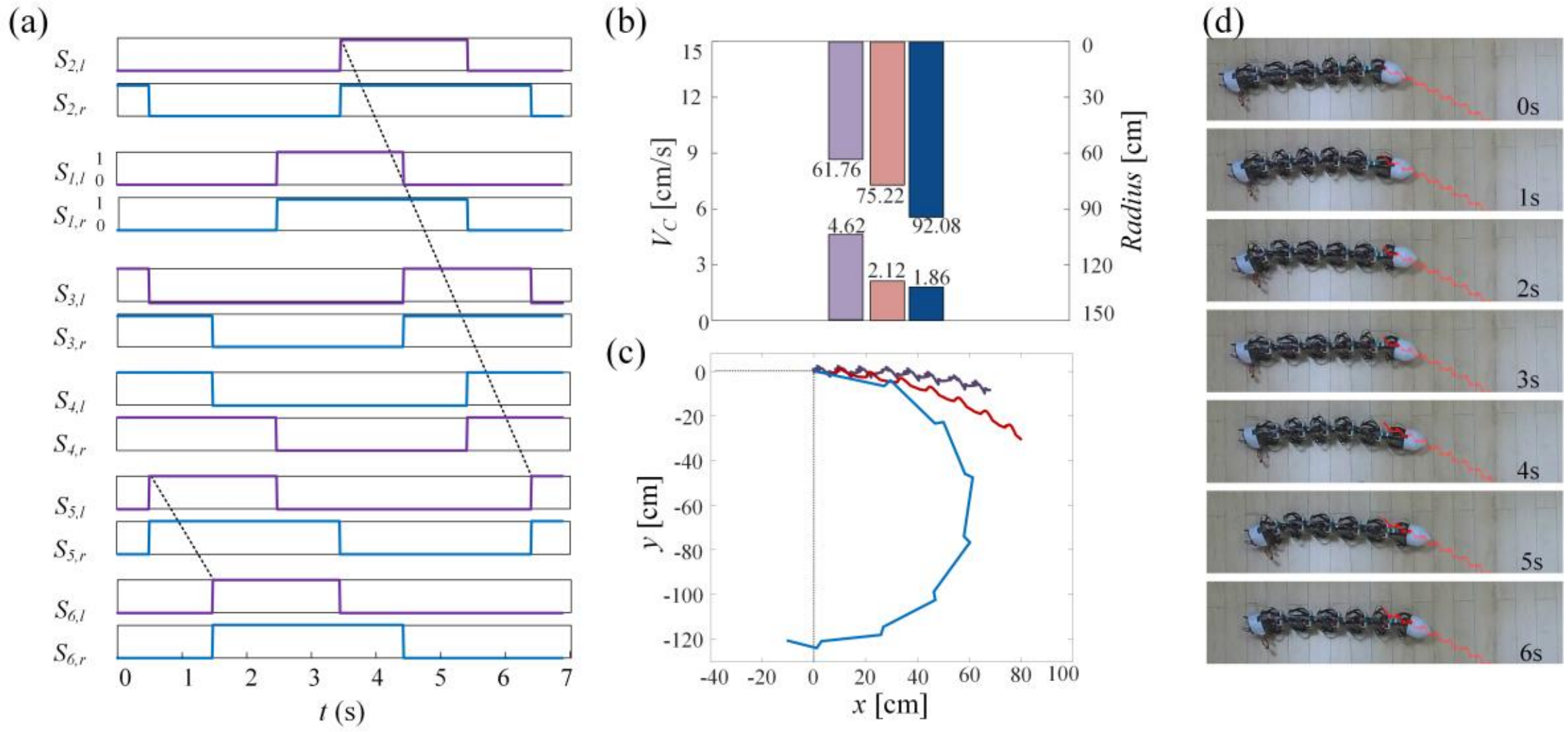

**FIG. 4.** Actuation signals and experiment of the earthworm-like circular locomotion. (a) Actuation signals of the 6 robot segments in a locomotion cycle. (b) Comparison of the kinematic, dynamic, and experimental average velocities and radius of the circular locomotion. (c) Displacement-time histories of the robot head under the circular gait. (d) Snapshots of the robot during the circular locomotion.

The simulations based on the dynamic model are shown to agree more with the experimental results than the kinematic model, as revealed in Fig. 2 (b), (c), Fig. 3 (b), (c)and Fig. 4 (b), (c). The kinematic model, on the other hand, deviates significantly from experimental results in predicting the slope of the sidewinding locomotion trajectory and the velocity of the circular locomotion. In contrast, the dynamic model gives reasonable predictions for these parameters. This is further supported by the displacement-time histories of the robot head, where the dynamic simulations' trajectories match experimental trajectories better than those of the kinematic simulations. However, although the dynamic model improves predictions by accounting for inertial and frictional forces, some deviations between the predicted and experimental results still exist, such as higher average velocities and larger predicted radii of trajectories in worm-like rectilinear locomotion. There are several possible explanations for these differences. First, the assumption of ideal actuation may not be accurate, meaning that the worm-like segments cannot deform as intended. Second, although the current dynamic model has incorporated forward, backward, and lateral friction, actual planar locomotion involves variable friction between the robot segment and the ground in different directions, with inconsistent friction coefficients at different ground locations, and floor gaps that can impact robot motion. Third, sideslips have an impact on robot locomotion, as in the case of sidewinding locomotion, where internal forces cause the head segment to be pulled back slightly by the posterior segments, resulting in negative displacements in the y-direction, or circular locomotion, where sideslip increases the radius of the trajectory. Fourth, the worm-like locomotion assumes that snake-like joints remain in the neutral position, but slight rotations may occur during the experiment due to joint torques. Finally, manual fabrication errors can result in a non-standard isosceles trapezoid shape when one side of the worm-like module is contracted.

## 4 Conclusion

A comprehensive analysis of the dynamics of a metamer-like earthworm robot capable of planar motion is presented in this work. The model takes into account the complex interactions between the robot's deformable body and the forces acting on it. The verification of the model through simulations shows that the dynamic model agrees more with experimental results than the kinematic model. However, there are still deviations between the predicted and experimental results, indicating that further research is needed. In conclusion, the proposed dynamic model for the earthworm-like robot's planar motion represents a significant advancement in the field of metameric robotics. This work is a crucial step towards unlocking the full potential of soft-bodied robots and advancing the field of Metameric robotics.

## Reference

[1] Webster RJ, Jones BA. Design and Kinematic Modeling of Constant Curvature Continuum Robots: A Review. Int J Rob Res 2010;29:1661–83.

[2] Shabalina K, Sagitov A, Magid E, IEEE. Comparative Analysis of Mobile Robot Wheels Design. 2018 11TH Int Conf Dev ESYSTEMS Eng (DESE 2018) 2018:175–9.

[3] Zhang Y, Yang D, Yan P, Zhou P, Zou J, Gu G. Inchworm Inspired Multimodal Soft Robots With Crawling, Climbing, and Transitioning Locomotion. IEEE Trans Robot 2022;38:1806–19.

[4] Jia Y, Ma S. A Decentralized Bayesian Approach for Snake Robot Control. IEEE Robot Autom Lett 2021;6:6955–60.

[5] Guo, Xian, Shugen Ma, Bin Li YF. A Novel Serpentine Gait Generation Method for Snakelike Robots Based on Geometry Mechanics. IEEE/ASME Trans Mechatronics 2018;23:1249–58.

[6] Jin YL, Ren J, Feng WB, Li JJ, Wang BR. Gait analysis of an inchworm-like robot climbing on curved surface and CPG-based planning. Binggong Xuebao/Acta Armamentarii 2016;37:1104–10.

[7] Q.Y. Tan Y.K. Wang C.N. Song ZLWCG. A SMA Actuated Earthworm-Like Robot 2011:619–24.

[8] Fang H, Wang C, Li S, Xu J, Wang KW. Design and experimental gait analysis of a multi-segment in-pipe robot inspired by earthworm’s peristaltic locomotion. Bioinspiration, Biomimetics, Bioreplication 2014, vol. 9055, 2014, p. 90550H.

[9] Boxerbaum A, Chiel H. Softworm: A Soft, Biologically Inspired Worm-Like Robot.

Neurosci Abstr 2009;315:44106.
[10] Xiong Zhan, Jian Xu HF. In-plane gait planning for earthworm-like metameric robots using genetic algorithm. Bioinspir Biomim 2018:0–5.
[11] Fang H, Li S, Wang KW, Xu J. Phase coordination and phase-velocity relationship in metameric robot locomotion. Bioinspiration and Biomimetics 2015;10:066006.
[12] E.Garrey W, A.R.Moore. Peristalsis and coordination in the earthworm 1915.
[13] Edwards CA, Bohlen PJ. Biology and Ecology of Earthworms. vol. 3. London: Chapman & Hall; 1996.
[14] Tang Z, Lu J, Wang Z, Chen W, Feng H. Design of a new air pressure perception multi-cavity pneumatic-driven earthworm-like soft robot. Auton Robots 2020;44:267–79.
[15] Matsushita K, Ikeda M, Or K, Niiyama R, Kuniyoshi Y, IEEE. An Actuation System using a Hydrostatic Skeleton and a Shape Memory Alloy for Earthworm-like Soft Robots. 2022 IEEE/SICE Int Symp Syst Integr (SII 2022) 2022:47–52.
[16] Tang Z, Lu J, Wang Z, Ma G, Chen W, Feng H. Development of a New Multi-cavity Pneumatic-driven Earthworm-like Soft Robot. Robotica 2020;38:2290–304.
[17] Zarrouk D, Shoham M. Analysis and design of one degree of freedom worm robots for locomotion on rigid and compliant terrain. J Mech Des Trans ASME 2012;134:1–9.
[18] Zhou X, Teng Y, Li X. Development of a new pneumatic-driven earthworm-like soft robot. M2VIP 2016 - Proc 23rd Int Conf Mechatronics Mach Vis Pract 2017:1–5.
[19] Fang H, Li S, Wang KW, Xu J. A comprehensive study on the locomotion characteristics of a metameric earthworm-like robot: Part A: Modeling and gait generation. Multibody Syst Dyn 2015;34:391–413.
[20] Marvi H, Bridges J, Hu DL. Snakes mimic earthworms: propulsion using rectilinear travelling waves. R Soc 2013.
[21] Nakamura T, Kato T, Iwanaga T, Muranaka Y. Peristaltic Crawling Robot Based on the Locomotion Mechanism of Earthworms. IFAC Proc Vol 2006.
[22] Schwebke S, Behn C. Worm-like robotic systems: Generation, analysis and shift of gaits using adaptive control. Artif Intell Res 2012;13:743–53.
[23] Daltorio KA, Boxerbaum AS, Horchler AD, Shaw KM, Chiel HJ, Quinn RD. Efficient worm-like locomotion: Slip and control of soft-bodied peristaltic robots. Bioinspiration and Biomimetics 2013;8.
[24] Aydin YO, Molnar JL, Goldman DI, Hammond FL. Design of a soft robophysical earthworm model. 2018 IEEE Int Conf Soft Robot RoboSoft 2018 2018:83–7.
[25] Zhan X, Fang H, Xu J, Wang KW. Planar locomotion of earthworm-like metameric robots. Int J Rob Res 2019;38:1751–74.
[26] Shugen. Analysis of creeping locomotion of a snake-like robot. Adv Robot 2001.
[27] Liljebäck P, Pettersen KY, Stavdahl Ø, Gravdahl JT. Snake Robots - Modelling, Mechatronics, and Control. Springer Publ Company, Inc 2012.